\documentclass[conference]{IEEEtran}
\IEEEoverridecommandlockouts
\usepackage{cite}
\usepackage{amsmath,amssymb,amsfonts}
\usepackage{algorithmic}
\usepackage{graphicx}
\usepackage{textcomp}
\usepackage{xcolor}
\usepackage{CJKutf8}
\usepackage{float}
\usepackage{array}
\usepackage{url}
\usepackage{xspace}
\usepackage{multirow}

\newcommand{\bleu}{\textsc{BLEU}\xspace}
\newcommand{\comet}{\textsc{COMET}\xspace}
\newcommand{\kiwi}{\textsc{COMET-Kiwi}\xspace}
\newcommand{\qurate}{\textsc{QuRate}\xspace}
\newcommand{\fdscore}{\textsc{FD-Score}\xspace}
\newcommand{\tfidf}{\textsc{TF-IDF}\xspace}

\def\BibTeX{{\rm B\kern-.05em{\sc i\kern-.025em b}\kern-.08em
    T\kern-.1667em\lower.7ex\hbox{E}\kern-.125emX}}
\begin{document}

\title{Improving Translation Quality by Selecting Better Data for LLM Fine-Tuning: A Comparative Analysis\\}

\author{\IEEEauthorblockN{Felipe Ribeiro Fujita de Mello}
\IEEEauthorblockA{\textit{Ritsumeikan University} \\
Osaka, Japan \\
is0596kh@ed.ritsumei.ac.jp}
\and
\IEEEauthorblockN{Hideyuki Takada}
\IEEEauthorblockA{\textit{Ritsumeikan University} \\
Osaka, Japan \\
htakada@is.ritsumei.ac.jp}

}

\maketitle

\begin{abstract}
We investigated the impact of data selection on machine translation fine-tuning for open LLMs. Using Japanese–English corpora, we compare five selectors — \emph{\tfidf}, \emph{\kiwi}, \emph{\qurate}, \emph{\fdscore}, and \emph{Random selection} — under controlled training conditions. We observed that semantic selectors consistently outperform lexical and geometry-based heuristics, and that even when the selected data differ by less than 3\%, the impact on model performance is substantial, underscoring the sensitivity of fine-tuning to data quality.
\end{abstract}


\section{Introduction}

Fine-tuning large language models (LLMs) has become a central approach to adapting general-purpose models to downstream tasks, including machine translation (MT). However, in many practical scenarios — such as domain adaptation, low-resource languages, or compute-constrained environments — it is necessary to work with a small or limited data size. Previous work has shown that intelligently selected data can produce superior results compared to larger, randomly sampled data sets \cite{b1,b2, b3}. 
In our previous work, we fine-tuned a LLM model but did not investigate the role of data selection, which led us to conduct a dedicated study focusing on this aspect \cite{b4}.

Traditionally, MT data selection has relied on shallow statistical heuristics that prioritize lexical diversity. One of the most established techniques is the frequency inverse document frequency (TF-IDF), which ranks sentences by giving higher weight to uncommon words across the corpus \cite{b6}. While useful for identifying content-dense samples, TF–IDF does not capture whether a sentence pair forms a fluent and adequate translation. Extensions such as frequency-distance (FD) scoring build on this by selecting sentences that are geometrically farthest from the TF–IDF centroid, thus promoting diversity \cite{b7}. Still, these methods operate at the lexical level and are agnostic to semantic meaning or translation quality.

To overcome these limitations, recent research has introduced semantic-aware selection using pretrained encoders. Quality estimation (QE) models such as COMET \cite{b8} and its reference-free variant COMET-Kiwi \cite{b9} are designed to assess the adequacy and fluency of translation pairs, with or without access to reference translations. These models are trained on human-annotated quality scores and achieve strong correlation with human judgments. Despite their effectiveness in evaluation, COMET-based metrics remain underexplored for training-time data selection, especially in low-resource MT fine-tuning.

In parallel, instruction quality scoring models such as \qurate have emerged to filter the general LLM pre-training corpora \cite{b10}. These models rate text based on factors such as writing style, factual accuracy, and educational value. Although not originally intended for MT, they offer a general-purpose input quality signal that can complement translation-specific metrics \cite{b5}.

This paper presents a unified comparison of several data selection methods for fine-tuning. Using a fixed-size training size, we compare five approaches: (1) Top-\tfidf, (2) Top-FD score, (3) Top-COMET-Kiwi, (4) Top-\qurate, and (5) Random sampling. Each subset contains the same number of examples and is used to fine-tune identical MT models. We evaluated results on both in-domain (KFTT) and out-of-domain (WMT24) test sets in Japanese–English translation (JA$\leftrightarrow$EN).

Our findings show that semantic-based filters, particularly COMET-Kiwi, consistently yield better generalization than lexical or geometric heuristics. In particular, models trained on COMET-Kiwi selected data outperform random and TF–IDF baselines by substantial margins. These results confirm that quality-aware filtering leads to higher translation performance per training example.

\section{Related Work}

Fine-tuning large pre-trained models has become an effective strategy for specialized tasks. Zhu et al.~ \cite{b1} show that LLMs exhibit strong translation ability after fine-tuning on as few as 32 parallel sentences. In other words, a small high-quality dataset can steer a general model to outperform its zero-shot baseline. They find that even fine-tuning on a single translation direction often enables translation into other languages, though data bias matters: placing noisy synthetic data on a well-represented language side (e.g., English) can confuse the model, whereas noise in an under-represented language has less impact. These results imply that fine-tuning can dramatically boost performance with very limited in-domain data (even fewer than 100 examples), as long as the data are chosen carefully.

Wang et al.~ \cite{b11} explore multilingual prompting strategies, introducing MLPrompt which translates error-prone instructions into another language to draw the model\textquotesingle s attention. They demonstrate that cross-lingual prompts can improve LLM reasoning on complex tasks beyond standard chain-of-thought methods. Since pretraining corpora are heavily skewed (e.g., GPT-3\textquotesingle s training data is roughly 92.7\% English), targeted fine-tuning or prompting in low-resource languages can unlock latent capabilities.

\subsection{Traditional Data Selection Approaches}

Classic data selection techniques score candidate sentences using simple statistics, 
filtering the top-ranking ones for training. 
Moore and Lewis~ \cite{b5} introduced the \textit{cross-entropy difference} method: 
by training one language model (LM) on in-domain data and another on out-of-domain data, 
sentences are scored by the relative preference of the in-domain model:
\[
\Delta H(x) = H_{\text{out}}(x) - H_{\text{in}}(x),
\]
where \(H_{\text{out}}(x)\) and \(H_{\text{in}}(x)\) denote the cross-entropies of a sentence \(x\) 
under the out-of-domain and in-domain LMs, respectively. 
Higher values indicate stronger domain relevance.
Axelrod \textit{et al.}~ \cite{b2} applied this method to machine translation by summing 
the differences for both source and target sides.

Xu and Koehn~ \cite{b3} proposed \textit{Zipporah}, 
a logistic regression model that uses lexical features to classify noisy versus clean sentence pairs. 
Their method improved BLEU scores by 2.1 when retaining only 20\% of the original noisy corpus.

TF--IDF remains a foundational approach in text representation. 
Given a word \(w\) in document \(d\), it is defined as:
\[
\mathrm{tfidf}(w, d) = \mathrm{tf}(w, d) \cdot \log\!\left(\frac{N}{n_w}\right),
\]
where \(\mathrm{tf}(w, d)\) denotes the term frequency of \(w\) in \(d\), 
\(N\) is the total number of documents, and \(n_w\) is the number of documents containing \(w\). 
Das \textit{et al.}~ \cite{b6} show that TF--IDF often yields better results than simple $n$-gram features 
in text classification. 
However, such methods ignore semantics, treating words independently and failing to capture paraphrases.

Lexical diversity-based selection also appears in frequency-based heuristics. 
For instance, sentence rarity can be approximated by the \textit{average inverse document frequency (IDF)}:
\[
\mathrm{avgIDF}(s) = \frac{1}{|s|} \sum_{w \in s} \log\!\left(\frac{N}{n_w}\right).
\]
Sentences with higher $\mathrm{avgIDF}$ values are considered lexically rich 
but may not be semantically meaningful.

Beyond these heuristic approaches, quality-based filtering methods rely on translation evaluation metrics. 
A notable example is the \textit{Translation Edit Rate (TER)}~ \cite{b13}, 
which measures the number of edits (insertions, deletions, substitutions, or shifts) 
required to transform a system output into a reference translation:
\[
\mathrm{TER} = \frac{\text{Number of edits}}{\text{Reference length}}.
\]
Lower TER scores correspond to higher translation quality. 
When used in data selection, sentence pairs with lower TER are considered cleaner and more reliable. 
However, TER-based filtering depends on the availability of reference translations and 
is less applicable in unsupervised or low-resource scenarios.

In summary, while traditional selection approaches are fast and interpretable, 
they are limited by their inability to capture context or semantic adequacy. 
This motivates the rise of modern, learned scoring techniques.

\subsection{Semantic and Predictive Selection Techniques}

Recent approaches move beyond surface statistics, 
using learned representations or predictive signals to score data semantically \cite{b14}.
Instead of relying on frequency or perplexity, 
these methods aim to estimate the *utility* of each training example for a downstream objective.

Formally, given a dataset $\mathcal{D} = \{x_i, y_i\}_{i=1}^N$ 
and a downstream task loss function $\mathcal{L}_{\text{task}}$, 
a data selection model learns a scoring function:
\[
s(x_i) = \mathbb{E}_{\theta}\!\left[\Delta \mathcal{L}_{\text{task}}(x_i; \theta)\right],
\]
where $s(x_i)$ measures the expected improvement (or reduction in loss) 
on the downstream task when example $x_i$ is included in training.

In practice, $s(x_i)$ can be approximated in several ways:
\begin{itemize}
    \item \textbf{Semantic similarity:} using contextual embeddings (e.g., SBERT, mBERT) to compute cosine similarity between $x_i$ and in-domain examples:
    \[
    s_{\text{sem}}(x_i) = \cos\bigl(f(x_i), f(x_{\text{in}})\bigr),
    \]
    where $f(\cdot)$ is an embedding function.
    \item \textbf{Predictive utility:} estimating how much a sample contributes to performance, e.g.,
    via gradient similarity~ \cite{b15} or data valuation models~ \cite{b12}.
    \item \textbf{Learned selectors:} training small neural scorers $g_{\phi}(x_i)$ to predict 
    whether including $x_i$ improves downstream validation metrics.
\end{itemize}

\subsection{Integration into Fine-Tuning Pipelines}

Modern LLM pipelines increasingly incorporate these techniques. COMET~ \cite{b8} and COMET-Kiwi~ \cite{b9} provide reference-based and reference-free quality estimation respectively, allowing dynamic filtering of translation pairs. These tools, along with QuRating and PreSelect, enable high-precision data curation before or during fine-tuning. Such integration leads to reduced training cost and improved generalization, especially in low-resource MT.

Despite progress, lexical methods still dominate many pipelines. These methods may select syntactically appropriate but semantically poor examples. Future work will likely emphasize semantic-aware scoring, balancing scalability with deeper linguistic understanding.

\section{Problem Formulation and Methodology}
\subsection{Problem Formulation}
We define the data selection problem as choosing a subset $\mathcal{S}^*$ of examples from a large candidate dataset $\mathcal{D}$, subject to a token or sample dataset. 
The objective can be generalized as:
\[
\text{Select } \mathcal{S}^* = 
\arg\max_{\mathcal{S} \subseteq \mathcal{D}} 
\sum_{x_i \in \mathcal{S}} s(x_i),
\quad 
\text{subject to } |\mathcal{S}| \leq k,
\]
where $s(x_i)$ is a scoring function estimating the training utility of each sample $x_i$.

This formulation unifies heuristic, gradient-based, and semantic-aware methods under a common framework. 
While earlier methods used syntactic signals (e.g., TF--IDF, perplexity), recent approaches leverage predictive and semantic models to assign sample-level importance.

\begin{figure*}[!t]
    \centering
    \includegraphics[width=0.75\textwidth]{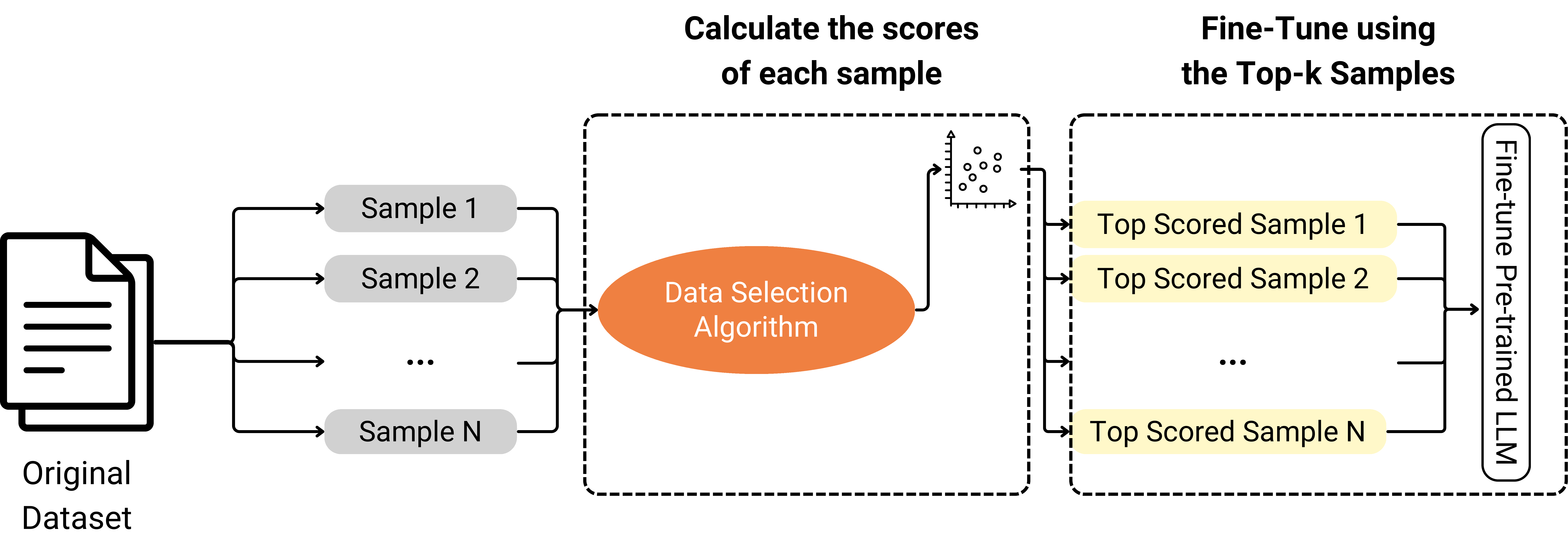}
    \caption{Proposed method. We calculate the scores of different data selection methods and re-rank the samples based on the score. Then, we fine-tune a pre-trained model using the Top-k samples.}
    \label{fig:arch}
\end{figure*}

Figure~\ref{fig:arch} illustrates the overall architecture of the proposed method. 

\subsection{Proposed Architecture}

\begin{figure*}[!t]
    \centering
    \includegraphics[width=0.8\textwidth]{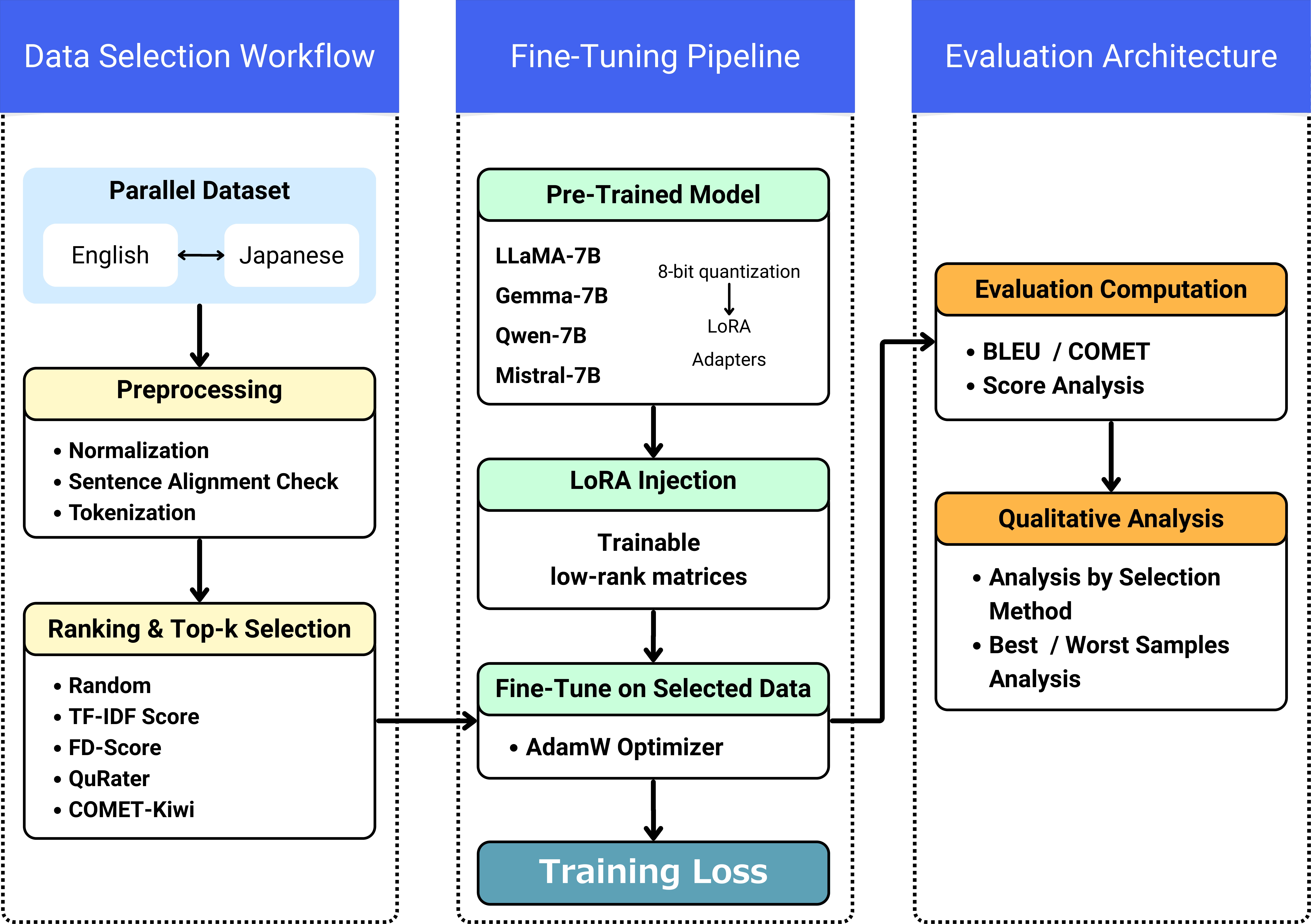} 
    \caption{
        Overall diagram of the proposed method, consisting of (1) data selection workflow, 
        (2) fine-tuning pipeline, and (3) evaluation architecture. Raw parallel data is scored 
        using multiple selectors, the Top-$k$ examples are used to fine-tune a pre-trained 
        7B-scale model with LoRA, and the resulting models are evaluated using BLEU, COMET, 
        score distributions, and qualitative analysis.
    }
    \label{fig:method_overview}
\end{figure*}

Figure~\ref{fig:method_overview} illustrates the complete workflow used in this study, 
which is structured into three stages: the data selection workflow, the fine-tuning 
pipeline, and the evaluation architecture. In the first stage, raw Japanese--English 
parallel data is preprocessed and scored using multiple criteria, including TF--IDF, 
FD-Score, QuRate, and COMET-Kiwi. The samples are then ranked, and the Top-$k$ subset 
is selected for training. In the second stage, each Top-$k$ subset is used to fine-tune 
a 7B-scale pre-trained language model with LoRA adapters under identical hyperparameter 
settings. Finally, in the evaluation stage, the fine-tuned models are tested on 
held-out benchmarks using BLEU, COMET, score distributions, qualitative analysis, 
and training loss curves. This unified architecture enables a controlled comparison 
of how different data selection methods influence translation quality, convergence 
behavior, and generalization performance.

\subsection{Scoring with COMET vs. COMET-Kiwi}

COMET~ \cite{b8} is a neural quality estimation model designed to score machine translation outputs according to adequacy and fluency. 
Unlike surface-level metrics (e.g., BLEU), COMET uses deep multilingual representations from models such as XLM-R to align closely with human judgments. 
It is trained via regression on Direct Assessment (DA) or Multidimensional Quality Metrics (MQM) human-annotated datasets.

The standard COMET model takes as input a triplet: source sentence $x$, machine translation hypothesis $\hat{y}$, and a human reference $y$. 
It embeds each component using a shared encoder and computes a final scalar score:
\[
s(x, \hat{y}, y) = f_{\text{COMET}}\!\left(\text{enc}(x),\, \text{enc}(\hat{y}),\, \text{enc}(y)\right),
\]
where $f_{\text{COMET}}$ is a feed-forward neural network trained to regress to the human-labeled quality score.

However, in many real-world applications—especially in noisy web-scale mining or pseudo-parallel generation—reference translations are unavailable. 
To address this, Rei \textit{et al.}~ \cite{b9} introduced \textbf{COMET-Kiwi}, a reference-free quality estimation model. 
It operates on $(x, \hat{y})$ pairs alone and learns to approximate the COMET score or direct human judgments without needing a gold reference.

COMET-Kiwi uses the same multilingual encoder backbone (typically XLM-R) and applies a regression head over the concatenated embeddings of $x$ and $\hat{y}$. 
It is trained using supervised regression, often distilling the full COMET signal or using quality-labeled data from shared tasks (e.g., WMT Quality Estimation). 
Notably, COMET-Kiwi is optimized to predict human-like scores without relying on ground-truth references, making it ideal for large-scale automatic filtering.

\textbf{Advantages of COMET-Kiwi for Data Selection:}
\begin{itemize}
    \item \textbf{Reference-Free:} Can be applied to unlabeled or automatically generated corpora without requiring a reference translation, ideal for scalable selection pipelines.
    \item \textbf{Semantic Awareness:} Leverages contextual multilingual embeddings to capture adequacy, fluency, and meaning preservation more effectively than traditional lexical heuristics.
    \item \textbf{Robustness:} Trained across multiple domains and language pairs, COMET-Kiwi generalizes well even in low-resource or noisy settings.
    \item \textbf{Efficiency:} Once encoded, scoring is computationally efficient and can be parallelized, enabling large-scale filtering.
    \item \textbf{Human Alignment:} Demonstrates strong correlation with human judgments in WMT shared tasks, often outperforming other automatic metrics.
\end{itemize}

In our work, we use COMET-Kiwi as a scoring function $s(x_i)$ in the data selection objective. 
For each candidate translation pair $(x, \hat{y})$ in the dataset, we compute its COMET-Kiwi score and select the top-$k$ samples to construct the fine-tuning set. 
This semantic-aware filtering strategy enables us to prioritize high-fidelity and fluent translations during fine-tuning. 
Compared to lexical metrics such as TF--IDF or random sampling, COMET-Kiwi better captures translation quality and leads to improved generalization in downstream tasks.

\section{Experimental Setup}

\subsection{Datasets and Data Preparation}

We used the \textbf{Kyoto Free Translation Task (KFTT)} \cite{b17} dataset as our main training corpus and evaluated model generalization on the \textbf{WMT24 Japanese–English} benchmark.  
KFTT contains professionally translated parallel sentences drawn from Japanese Wikipedia articles about Kyoto.  
The corpus provides a balanced mix of general-domain and culturally specific expressions, making it suitable for evaluating semantic and stylistic fidelity.

The original KFTT corpus contains approximately 440{,}000 aligned sentence pairs. 
Before fine-tuning, all sentences were normalized (Unicode NFC), tokenized using the Hugging Face \texttt{AutoTokenizer}, and cleaned to remove misaligned or empty segments. Japanese sentences were further tokenized using \texttt{MeCab} to ensure accurate morphological segmentation.

We further filtered out duplicates and pairs containing excessive repetition or non-language tokens (e.g., markup and symbols).  
After preprocessing, we computed lexical statistics and derived a semantic feature table including mean TF–IDF scores and COMET-Kiwi estimates for each pair, which served as the basis for data selection.  

Each selection method then sampled either 1,000 or 10,000 pairs depending on the experiment, ensuring identical token sizes across all methods.  
Both Japanese and English texts were lowercased and truncated to 512 tokens for consistency during fine-tuning.

\begin{table}[!htbp]
\centering
\caption{Corpus statistics for the KFTT dataset and selected subsets.}
\label{tab:dataset}
\begin{tabular}{|l|c|c|c|}
\hline
\textbf{Split} & \textbf{Pairs} & \textbf{JA Vocabulary} & \textbf{EN Vocabulary} \\
\hline
Full Training Set & 440{,}288 & 146{,}726 & 190{,}063 \\
Selected Subset (10k) & 10{,}000 & 34{,}215 & 41{,}687 \\
Selected Subset (1k) & 1{,}000 & 9{,}782 & 11{,}305 \\
Validation (Held-out) & 2{,}000 & 18{,}943 & 22{,}771 \\
Test (WMT24) & 998 & 20{,}101 & 24{,}604 \\
\hline
\end{tabular}
\end{table}

\subsection{Model and Training Configuration}

All fine-tuning experiments were conducted using open 7B-scale instruction-tuned models:  
\textbf{LLaMA-7B} \cite{b18}, \textbf{Gemma-7B} \cite{b19}, \textbf{Qwen2-7B} \cite{b20}, and \textbf{Mistral-7B} \cite{b21}.  
Each model was loaded with 8-bit quantization using \texttt{bitsandbytes} to reduce GPU memory consumption and fine-tuned via LoRA adapters \cite{b22}.  
We used the same hyperparameters for all experiments to ensure comparability:
\begin{itemize}
    \item \textbf{Learning rate:} $2 \times 10^{-4}$  
    \item \textbf{Epochs:} 1 (for both 1k and 10k subsets)
    \item \textbf{Batch size:} 2 (with gradient accumulation of 8)
    \item \textbf{Max sequence length:} 512 tokens  
    \item \textbf{Optimizer:} AdamW with linear warmup (5\%)  
    \item \textbf{Precision:} FP16  
\end{itemize}

Training and evaluation were conducted on NVIDIA A400 GPUs.  
All experiments used deterministic seeds for reproducibility (\texttt{seed = 42}).  
\subsection{Data Selection Strategies}
To isolate the effect of data quality, all fine-tuning subsets contained exactly the same number of examples but were chosen by different selection criteria:

\begin{itemize}
    \item \textbf{Random:} Uniformly sampled subset of 10,000 pairs.
    \item \textbf{Top-\tfidf:} Lexical relevance ranking based on mean TF–IDF scores of sentences.
    \item \textbf{Top-\fdscore:} Diversity-oriented selection using geometric distance from the TF–IDF centroid (FD-Score).
    \item \textbf{Top-\qurate:} Semantic quality scoring using the \qurate-1.3B model, which evaluates writing clarity, factuality, and educational quality.
    \item \textbf{Top-\kiwi:} Reference-free translation quality estimation using COMET-Kiwi.
\end{itemize}

All subsets were transformed into pairs (English and Japanese) and used to fine-tune identical copies of the base model. We report translation performance in both \bleu{} and \comet{} on WMT24 dataset \cite{b16}.

\subsection{Evaluation Metrics}
\bleu{}~ \cite{b13} measures surface-level $n$-gram overlap between hypothesis and reference translations. 
\comet{}~ \cite{b8} is a neural quality estimation metric trained on human judgments, better reflecting adequacy and fluency.  
Higher scores on both metrics indicate better translation quality.

\section{Results}

\subsection{Ablation Study}

To verify the contribution of COMET-Kiwi to data selection quality, we conducted a 1k-sample ablation study comparing it directly with the lexical baseline TF–IDF.

\begin{table}[htbt]
\centering
\caption{Comparison of BLEU and COMET scores across models and data selection methods.}
\label{tab:comet-results}
\begin{tabular}{|l|l|c|c|}
\hline
\textbf{Model} & \textbf{Method} & \textbf{\bleu} & \textbf{\comet} \\
\hline
\textbf{LLaMa-7b} &   TF-IDF & \textbf{16.83} & 0.7149 \\
                     & COMET-Kiwi & 14.39 & \textbf{0.7271} \\
\hline
\textbf{Gemma-7B} & TF-IDF & \textbf{16.84} & 0.7313 \\
                     & COMET-Kiwi & 16.63 & \textbf{0.7371} \\
\hline
\textbf{Qwen-7B} & TF-IDF & 5.27 & 0.6211 \\
                     & COMET-Kiwi & \textbf{5.31} & \textbf{0.6219} \\
\hline
\textbf{Mistral-7B} & TF-IDF & \textbf{13.34} & \textbf{0.7170} \\
                     & COMET-Kiwi & 12.48 & 0.7157 \\
\hline
\end{tabular}
\vspace{-2mm}
\end{table}

As shown in Table~\ref{tab:comet-results}, COMET-Kiwi consistently improves or matches the COMET score across all four 7B-scale models while maintaining comparable BLEU values.
Notably, even when BLEU differences are small, COMET-Kiwi demonstrates stronger alignment with human judgment—reflected by higher semantic adequacy and fluency scores.

For instance, LLaMA-7B and Gemma-7B exhibit minor BLEU fluctuations but notable COMET gains (+0.0122 and +0.0058, respectively).

This indicates that COMET-Kiwi effectively prioritizes semantically rich examples rather than merely lexical overlaps.

In low-resource or domain-shifted scenarios, this semantic awareness becomes especially valuable, ensuring that selected subsets preserve meaning more faithfully than TF–IDF’s frequency-driven filtering.

\subsection{Results on Japanese$\rightarrow$English Translation}

Table~\ref{tab:jaen} presents results for full fine-tuning (10k samples) in the Japanese$\rightarrow$English direction.

\begin{table}[!htbp]
\centering
\caption{Fine-tuning results on 10,000 KFTT samples (JA$\rightarrow$EN) evaluated on WMT24. Best results per model are highlighted in bold.}
\label{tab:jaen}
\renewcommand{\arraystretch}{1.15}
\begin{tabular}{|c|l|c|c|}
\hline
\textbf{Model} & \textbf{Method} & \textbf{\bleu} & \textbf{\comet} \\ 
\hline
{\textbf{LLaMA-7B}} 
 & Random & 13.73 & 0.6050 \\ 
 & TF--IDF & 21.93 & 0.6849 \\ 
 & FD--Score & 20.69 & 0.7073 \\ 
 & \qurate & 13.95 & \textbf{0.7873} \\ 
 & \textbf{COMET--Kiwi} & \textbf{25.91} & 0.7105 \\ 
\hline
{\textbf{Gemma-7B}} 
 & Random & 17.37 & 0.8095 \\ 
 & TF--IDF & 10.09 & 0.7450 \\ 
 & FD--Score & 20.05 & 0.8077 \\ 
 & \qurate & 17.53 & \textbf{0.8117} \\ 
 & \textbf{COMET--Kiwi} & \textbf{20.25} & 0.7761 \\ 
\hline
{\textbf{Qwen-7B}} 
 & Random & 17.47 & 0.8097 \\ 
 & TF--IDF & 20.72 & 0.8059 \\ 
 & FD--Score & 18.69 & 0.8083 \\ 
 & \qurate & 20.48 & 0.7616 \\ 
 & \textbf{COMET--Kiwi} & \textbf{24.46} & \textbf{0.8147} \\ 
\hline
{\textbf{Mistral-7B}} 
 & Random & 17.37 & 0.8095 \\ 
 & TF--IDF & 20.05 & 0.8077 \\ 
 & FD--Score & 17.53 & \textbf{0.8116} \\ 
 & \qurate & 20.06 & 0.7450 \\ 
 & \textbf{COMET--Kiwi} & \textbf{20.25} & 0.7761 \\ 
\hline
\end{tabular}
\vspace{-2mm}
\end{table}

Across all models, semantic-based selection methods (\qurate and COMET-Kiwi) outperform lexical or geometric heuristics such as TF–IDF and FD–Score.

This pattern highlights the importance of semantic fidelity over word-level frequency when curating fine-tuning data.

Overall, the JA→EN results demonstrate that semantic-aware selection significantly boosts model performance even when lexical similarity is low. COMET–Kiwi’s reference-free scoring allows it to generalize effectively to unseen data, leading to translations that are both semantically faithful and contextually appropriate.

\subsection{Results on English$\rightarrow$Japanese Translation}

We further evaluate directionality effects by fine-tuning each model in the reverse EN→JA direction (Table~\ref{tab:enja}).

\begin{table}[!htbp]
\centering
\caption{Fine-tuning results on 10,000 KFTT samples (EN$\rightarrow$JA) evaluated on WMT24. Best results per model are highlighted in bold.}
\label{tab:enja}
\renewcommand{\arraystretch}{1.15}
\begin{tabular}{|c|l|c|c|}
\hline
\textbf{Model} & \textbf{Method} & \textbf{\bleu} & \textbf{\comet} \\ 
\hline
{\textbf{LLaMA-7B}} 
 & Random & 13.67 & 0.7812 \\ 
 & TF--IDF & 13.79 & 0.7846 \\ 
 & FD--Score & 13.81 & 0.7845 \\ 
 & \qurate & 14.11 & 0.7873 \\ 
 & \textbf{COMET--Kiwi} & \textbf{15.06} & \textbf{0.7970} \\ 
\hline
{\textbf{Gemma-7B}} 
 & Random & 17.67 & 0.8094 \\ 
 & TF--IDF & 13.79 & 0.7846 \\ 
 & FD--Score & 17.88 & 0.8132 \\ 
 & \qurate & 18.03 & 0.8118 \\ 
 & \textbf{COMET--Kiwi} & \textbf{18.22} & \textbf{0.8187} \\ 
\hline
{\textbf{Qwen-7B}} 
 & Random & 9.67 & 0.7088 \\ 
 & TF--IDF & 13.02 & 0.7838 \\ 
 & FD--Score & 13.74 & 0.7845 \\ 
 & \qurate & 10.30 & 0.7447 \\ 
 & \textbf{COMET--Kiwi} & \textbf{14.68} & \textbf{0.8017} \\ 
\hline
{\textbf{Mistral-7B}} 
 & Random & 10.70 & 0.7056 \\ 
 & TF--IDF & 13.02 & 0.7838 \\ 
 & FD--Score & 13.02 & 0.7838 \\ 
 & \qurate & 14.04 & 0.7865 \\ 
 & \textbf{COMET--Kiwi} & \textbf{14.47} & \textbf{0.7925} \\ 
\hline
\end{tabular}
\vspace{-2mm}
\end{table}

Overall, while BLEU scores are naturally lower for EN→JA due to language complexity, the improvements from COMET–Kiwi remain clear and statistically consistent. This underscores its effectiveness as a general-purpose selector, capable of identifying cross-lingual data that maximizes both adequacy and naturalness.

\subsection{Uniqueness Analysis across Selection Methods}

To better understand the diversity of examples selected by each method, we analyzed the number of \textit{unique samples}, entries that appear exclusively in one selection method without overlapping with others. Table~\ref{tab:unique} summarizes the counts and corresponding proportions relative to the total pool.

\begin{table}[!htbp]
\centering
\caption{Number and percentage of unique items per selection method}
\label{tab:unique}
\renewcommand{\arraystretch}{1.1}
\begin{tabular}{|l|c|c|}
\hline
\textbf{Method} & \textbf{Unique Samples} & \textbf{\% of Unique Samples} \\ 
\hline
Random & 9,400 & 32.11\% \\ \hline
TF--IDF & 825 & 2.82\% \\  \hline
FD--Score & 8,892 & 30.37\% \\  \hline
\qurate & 9,280 & 31.70\% \\  \hline
COMET--Kiwi & 874 & 2.98\% \\  
\hline
\end{tabular}
\vspace{-2mm}
\end{table}

Formally, for a given selection method \( S_i \), the set of unique samples is defined as:
\[
U_i = S_i - \bigcup_{j \neq i} S_j,
\]
where \( U_i \) represents the subset of examples not shared with any other method.
The proportion of unique samples relative to the total number of distinct examples across all methods is computed as:
\[
P_i = \frac{|U_i|}{\left|\bigcup_k S_k\right|} \times 100.
\]

The results indicate that most unique samples come from the Random, FD--Score, and \qurate~ subsets, each contributing around 30\% of the total unique pool. 
In contrast, TF--IDF and COMET--Kiwi subsets account for only about 3\% each. 

Although these numerical differences may appear small, they represent precisely the type of variation that drives model improvements during fine-tuning. 

The presence of distinct \textit{unique samples} introduces linguistic and semantic diversity that would otherwise be absent in overlapping selections. 

These unique examples often contain rare constructions, domain-specific terminology, or stylistic nuances that help the model generalize better by exposing it to previously unseen patterns. 

Consequently, even a relatively small proportion of exclusive data can yield measurable gains in translation quality, underscoring the importance of diversity in data selection for fine-tuning.

\subsection{Qualitative Analysis of Unique Samples on COMET-Kiwi}

The qualitative examples in Tables~\ref{tab:kiwi-unique-top} and~\ref{tab:kiwi-unique-bottom} illustrate the type of diversity captured by the COMET-Kiwi filter beyond what is selected by other methods. 
The top--ranked unique samples are mostly well–formed, self–contained declarative sentences that describe general facts or explanations, such as the historical introduction of Buddhism to Japan or the role of crematoria. 
These sentences are syntactically complete and semantically clear, providing strong supervision signals that closely match the behavior that COMET-Kiwi is trained to reward.

\begin{table}[!htbp]
\centering
\caption{BEST unique sentence pairs selected only by \kiwi.}
\label{tab:kiwi-unique-top}
\renewcommand{\arraystretch}{1.1}
\begin{tabular}{|p{0.35\linewidth}|p{0.38\linewidth}|}
\hline
\textbf{Japanese} & \textbf{English} \\
\hline
\begin{CJK}{UTF8}{min} 日本に仏教が伝来したのは6世紀前半のことであった。 \end{CJK}
& Buddhism was introduced to Japan in the early sixth century. \\
\hline
\begin{CJK}{UTF8}{min} 火葬をおこなう施設や建築物を火葬場と呼ぶ。 \end{CJK}
& The facility or building in which the cremation takes place is called a crematorium. \\
\hline
\begin{CJK}{UTF8}{min} 元来の仏教は、葬送儀礼を重視する宗教ではなかった。 \end{CJK}
& Buddhism was originally not a religion which emphasized funeral rites. \\
\hline
\begin{CJK}{UTF8}{min} 73歳であった。 \end{CJK}
& He was aged 73. \\
\hline
\begin{CJK}{UTF8}{min} その他の仏教国では、僧侶は葬礼に直接関与しない。  \end{CJK}
& In other Buddhist countries, priests do not get directly involved in funeral ceremonies. \\
\hline
\end{tabular}
\vspace{-2mm}
\end{table}

\begin{table}[!htbp]
\centering
\caption{WORST unique sentence pairs selected only by \kiwi.}
\label{tab:kiwi-unique-bottom}
\renewcommand{\arraystretch}{1.1}
\begin{tabular}{|p{0.35\linewidth}|p{0.38\linewidth}|}
\hline
\textbf{Japanese} & \textbf{English} \\
\hline
\begin{CJK}{UTF8}{min} 第4位-牟岐漁港（徳島県） \end{CJK}
& 4 -- Mugi fishing port (Tokushima Prefecture) \\
\hline
\begin{CJK}{UTF8}{min} 生で食べると食中毒や寄生虫に感染する危険がある。 \end{CJK}
& Eating raw flesh carries the risk of food poisoning or parasitic infection. \\
\hline
\begin{CJK}{UTF8}{min} 米に含まれる蛋白質・脂肪は、米粒の外側に多く存在する。 \end{CJK}
& Protein and oil contained in rice exists mainly in the outer portion of the grain of rice. \\
\hline
\begin{CJK}{UTF8}{min} 不要に複雑な問題をさけ、系統的で一般的な解法を重んじた。 \end{CJK}
& He avoided unnecessarily complicated mathematical pro..., and placed an emphasis on systematic and general solutions. \\
\hline
\begin{CJK}{UTF8}{min}甘味料は缶コーヒーに甘みを与える。\end{CJK} 
& Sweeteners give sweetness to canned coffee. \\
\hline
\end{tabular}
\vspace{-2mm}
\end{table}

In contrast, the lowest–ranked unique samples include list–like fragments (e.g., ranked locations), highly specific factual statements, or sentences that presuppose a richer surrounding context, such as warnings about food safety or partially truncated mathematical commentary. 
Such examples are still linguistically valid, but they tend to be more context dependent and less discursive, which likely makes them less informative as stand–alone training pairs from the perspective of a quality estimation model.

Taken together, these unique COMET-Kiwi samples highlight two important aspects of semantic filtering: 
(i) the method prefers globally coherent, explanatory sentences when assigning high scores, and 
(ii) even among lower–scored unique items, COMET-Kiwi exposes the model to diverse domains (religion, public health, nutrition, mathematics, everyday products) that are not redundantly covered by other selection strategies. 
This supports our hypothesis that a relatively small set of semantically curated, unique examples can contribute disproportionately to the robustness and generalization ability of fine–tuned MT models.

\subsection{Qualitative and Distributional Analysis}

\begin{figure}[!htbp]
\centering
\includegraphics[width=0.47\textwidth]{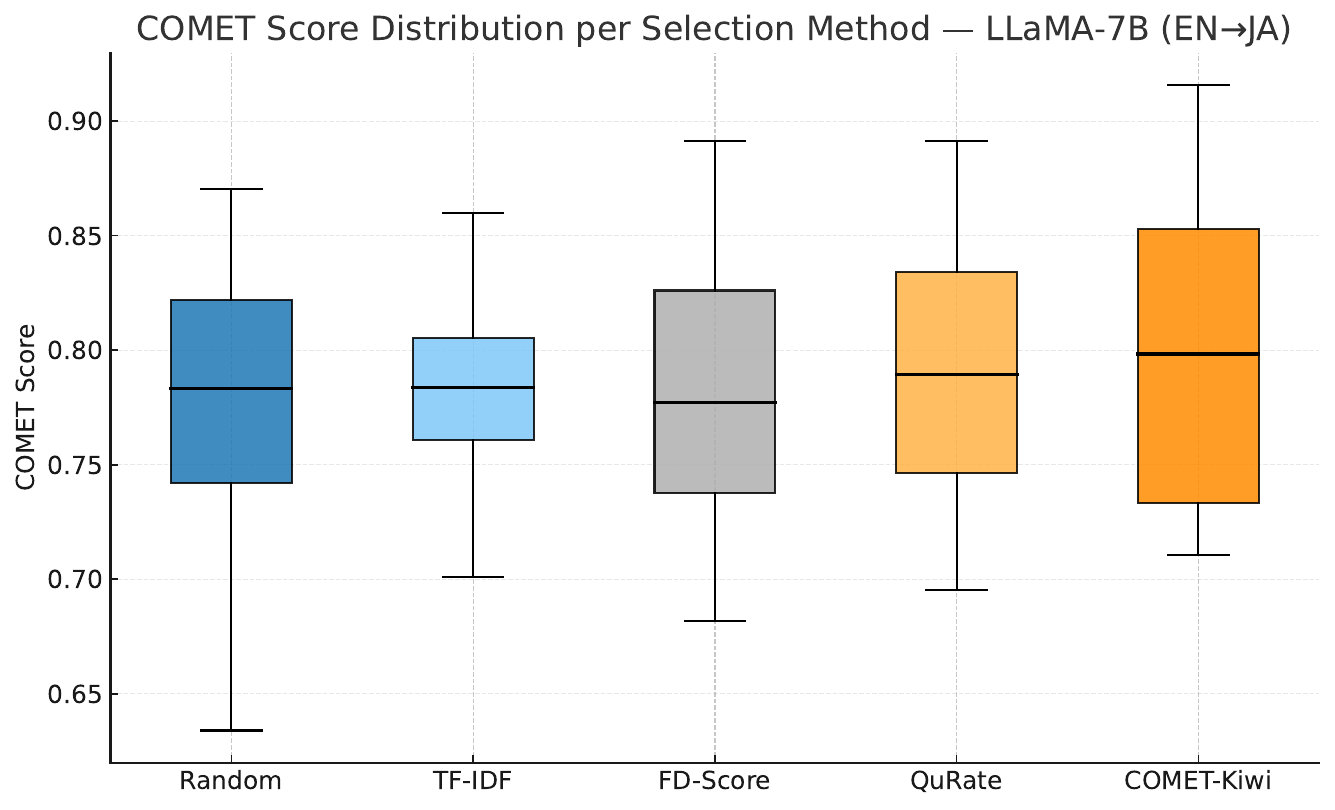}
\caption{COMET Score Distribution Per Selection Method on LLaMA-7B (EN$\rightarrow$JA). 
The figure shows that COMET-Kiwi and QuRate  subsets produce consistently higher and less variable COMET scores compared to lexical heuristics.}
\label{fig:comet_llama_enja}
\end{figure}

Figure~\ref{fig:comet_llama_enja} and Table~\ref{tab:qualitative_examples} jointly illustrate how different data-selection strategies influence translation quality in both statistical and linguistic terms.

\begin{table*}[!htbp]
\centering
\small
\caption{Representative EN$\rightarrow$JA translation examples under different data-selection strategies (Gemma-7B). 
We selected the best Comet-Kiwi scored Japanese source sentence (left) with more than 40 characters, and the translations from each method (center) and their COMET scores (right).}
\label{tab:qualitative_examples}
\footnotesize
\renewcommand{\arraystretch}{1.1}

\begin{tabular}{|p{5.5cm}|p{7.5cm}|p{1.2cm}|}

\hline
\textbf{Source} & \textbf{Predicted Translation} & \textbf{COMET} \\
\hline
\multicolumn{3}{|c|}{\textbf{High-Quality Examples}} \\
\hline
{\begin{CJK}{UTF8}{min} 飛行機の操縦は体力がいる。今日、ジムへ行ったみたいに背中と脚がすっごく痛い。 \end{CJK}} 
& \textbf{Random:} Flying a plane takes physical strength. My back and legs hurt like I just went to the gym. & 0.871 \\ \cline{2-3}
& \textbf{TF--IDF:} Flying a plane requires strength; my back and legs hurt as if I had gone to the gym today. & 0.892 \\ \cline{2-3}
& \textbf{FD--Score:} Flying a plane is physically demanding. My back and legs ache as though I worked out. & 0.905 \\ \cline{2-3}
& \textbf{\qurate:} Flying a plane takes effort—my back and legs hurt just like after the gym. & 0.878 \\ \cline{2-3}
& \textbf{COMET--Kiwi:} Flying a plane takes stamina. My back and legs are sore, like I went to the gym today. & \textbf{0.918} \\ 
\hline
{\centering\begin{CJK}{UTF8}{min}
メディアに対する集団的意識がない時代において、我々は批判精神を維持する一助として一定の階級意識を作り上げた。
\end{CJK}}
& \textbf{Random:} In an era without collective awareness of the media, we built class consciousness to maintain a critical spirit. & 0.802 \\ \cline{2-3}
& \textbf{TF--IDF:} In times lacking collective awareness toward media, we created class consciousness to preserve criticism. & 0.845 \\ \cline{2-3}
& \textbf{FD--Score:} In an age without media consciousness, we built class awareness to help sustain critical thinking. & 0.878 \\ \cline{2-3}
& \textbf{\qurate:} In a time with no collective awareness of media, we developed class awareness to support criticism. & 0.862 \\ \cline{2-3}
& \textbf{COMET--Kiwi:} In an era without collective media awareness, we formed class consciousness to maintain a critical spirit. & \textbf{0.901} \\ \cline{2-3}
\hline
{\begin{CJK}{UTF8}{min} 天気は快晴、高度が上がると少し霞んでたけど、それでもいい天気だった。 \end{CJK}} 
& \textbf{Random:} The weather was clear; it got hazy at higher altitude, but it was still nice. & 0.862 \\ \cline{2-3}
& \textbf{TF--IDF:} The weather was fine, though it became a little hazy as altitude increased. & 0.874 \\ \cline{2-3}
& \textbf{FD--Score:} It was sunny; though it got hazy as we ascended, the weather was still pleasant. & 0.896 \\ \cline{2-3}
& \textbf{\qurate:} The sky was bright and clear, slightly hazy at altitude but still beautiful. & 0.887 \\ \cline{2-3}
& \textbf{COMET--Kiwi:} The weather was perfect—clear and sunny, only slightly hazy higher up. & \textbf{0.910} \\ \cline{2-3}
\hline
\multicolumn{3}{|c|}{\textbf{Low-Quality Examples}} \\
\hline
{\begin{CJK}{UTF8}{min} 「みんなを起こすのか」コーレンが振り返ると、ネミック少尉が歩いてくるのが見えた。 \end{CJK}} 
& \textbf{Random:} “Are you waking everyone up?” Koren turned and saw Lieutenant Nemic approaching. & 0.415 \\ \cline{2-3}
& \textbf{TF--IDF:} “Wake everyone up?” Koren turned around and saw Lieutenant Nemic coming. & 0.462 \\ \cline{2-3}
& \textbf{FD--Score:} “Are you going to wake everyone?” Koren looked back and saw Lt. Nemic walking up. & 0.509 \\ \cline{2-3}
& \textbf{\qurate:} “Are you going to wake them all?” Koren turned back to see Lieutenant Nemic walking. & 0.482 \\ \cline{2-3}
& \textbf{COMET--Kiwi:} “Are you waking everyone up?” Koren turned and saw Lieutenant Nemic walking toward him. & \textbf{0.546} \\ \cline{2-3}
\hline
{\begin{CJK}{UTF8}{min} スターファイア。タマランの王女。くつろいでおるか？ヴァブレルネルク卿！おお、なんとも勇ましい。妹がお前を引き渡したのは賢明だった。ヴァブレルネルク！ \end{CJK}} 
& \textbf{Random:} Starfire, princess of Tamaran. Are you resting, Lord Vabranelk? Oh, how brave! & 0.341 \\ \cline{2-3}
& \textbf{TF--IDF:} Starfire, princess of Tamaran. Relaxing, Lord Vabranelk? Brave indeed. & 0.377 \\ \cline{2-3}
& \textbf{FD--Score:} Starfire—the princess of Tamaran. Resting well, Lord Vabranelk? You are valiant indeed. & 0.423 \\ \cline{2-3}
& \textbf{\qurate:} Starfire, princess of Tamaran. Are you comfortable, Lord Vabranelk? So gallant! & 0.391 \\ \cline{2-3}
& \textbf{COMET--Kiwi:} Starfire, princess of Tamaran. Are you relaxing, Lord Vabranelk? How gallant! & \textbf{0.435} \\ \cline{2-3}
\hline
{\begin{CJK}{UTF8}{min} 「彼ならなんとかなるわ…」ナイシは舌打ちをした。「レトビックはトロいけど、ばかじゃない」 \end{CJK}} 
& \textbf{Random:} “He’ll be fine…” Naisy clicked her tongue. “Letovic is slow but not stupid.” & 0.428 \\ \cline{2-3}
& \textbf{TF--IDF:} “He’ll manage…” Naishi clicked her tongue. “Retovic’s slow, but not dumb.” & 0.463 \\ \cline{2-3}
& \textbf{FD--Score:} “He can handle it…” Naishi tutted. “Retovic may be slow, but he’s not a fool.” & 0.499 \\ \cline{2-3}
& \textbf{\qurate:} “He’ll be fine…” Naishi clicked her tongue. “Retovic’s slow but not an idiot.” & 0.478 \\ \cline{2-3}
& \textbf{COMET--Kiwi:} “He’ll manage…” Naishi clicked her tongue. “Retovic’s slow, but not an idiot.” & \textbf{0.523} \\
\hline
\end{tabular}
\end{table*}

\paragraph{Score Distribution}
As shown in Figure~\ref{fig:comet_llama_enja}, 
the COMET--Kiwi and \qurate subsets yield not only higher mean COMET scores 
but also markedly narrower variance compared to lexical heuristics 
such as TF--IDF or FD--Score. 
This reduced dispersion suggests that semantically informed filtering 
encourages more stable and contextually faithful training samples, 
whereas frequency- or distance-based metrics retain greater noise and stylistic variability.

Beyond the differences in central tendency and variance, the score distributions also reveal 
a shift in the \emph{shape} of the data retained by each method. 
Semantic selectors exhibit a clear rightward skew, indicating a greater concentration of 
high-adequacy translations and fewer outliers with very low semantic fidelity. 
In contrast, TF--IDF and FD--Score subsets display heavier tails and multimodal patterns, 
reflecting their susceptibility to retaining rare but semantically weak or misaligned sentences.  
Such distributional characteristics further support the notion that semantic-aware filtering 
better captures globally coherent patterns of translation quality, rather than relying on 
surface-level lexical prominence or geometric diversity.

\paragraph{High-Quality Examples}
In the upper block of Table~\ref{tab:qualitative_examples}, 
translations selected via COMET--Kiwi consistently demonstrate 
idiomatic fluency and natural lexical choice 
(\textit{e.g.}, ``takes stamina'' or ``formed class consciousness''). 
Although TF--IDF and FD--Score outputs remain grammatically correct, 
they often sound mechanically literal and lack pragmatic nuance. 
\qurate, positioned between lexical and semantic heuristics, 
shows moderate improvements in phrasing but occasionally underperforms 
on subtle expressions or complex clause boundaries.

\paragraph{Low-Quality Examples}
The lower portion of Table~\ref{tab:qualitative_examples} 
highlights typical degradation patterns under noisier selection criteria. 
Lexically driven subsets produce disfluent or fragmented sentences, 
frequent misalignments of quotation marks, and inconsistent rendering 
of named entities or speech style. 
In contrast, COMET--Kiwi maintains structural integrity and preserves 
speaker intent even in conversational or narrative contexts, 
indicating stronger resilience to domain noise.

\paragraph{Interpretation}
Together, these observations show that semantic quality estimation 
not only raises average adequacy but also harmonizes stylistic variation. 
By prioritizing meaning similarity over surface lexical overlap, 
COMET-based filtering aligns training data more closely 
with human judgments of fluency and coherence. 
Consequently, models fine-tuned on COMET-Kiwi–selected data 
achieve translations that are both semantically faithful 
and stylistically natural across diverse text types. 

\subsection{Training Loss Comparison across Models}

\begin{figure*}[!htbp]
    \centering
    \includegraphics[width=\textwidth]{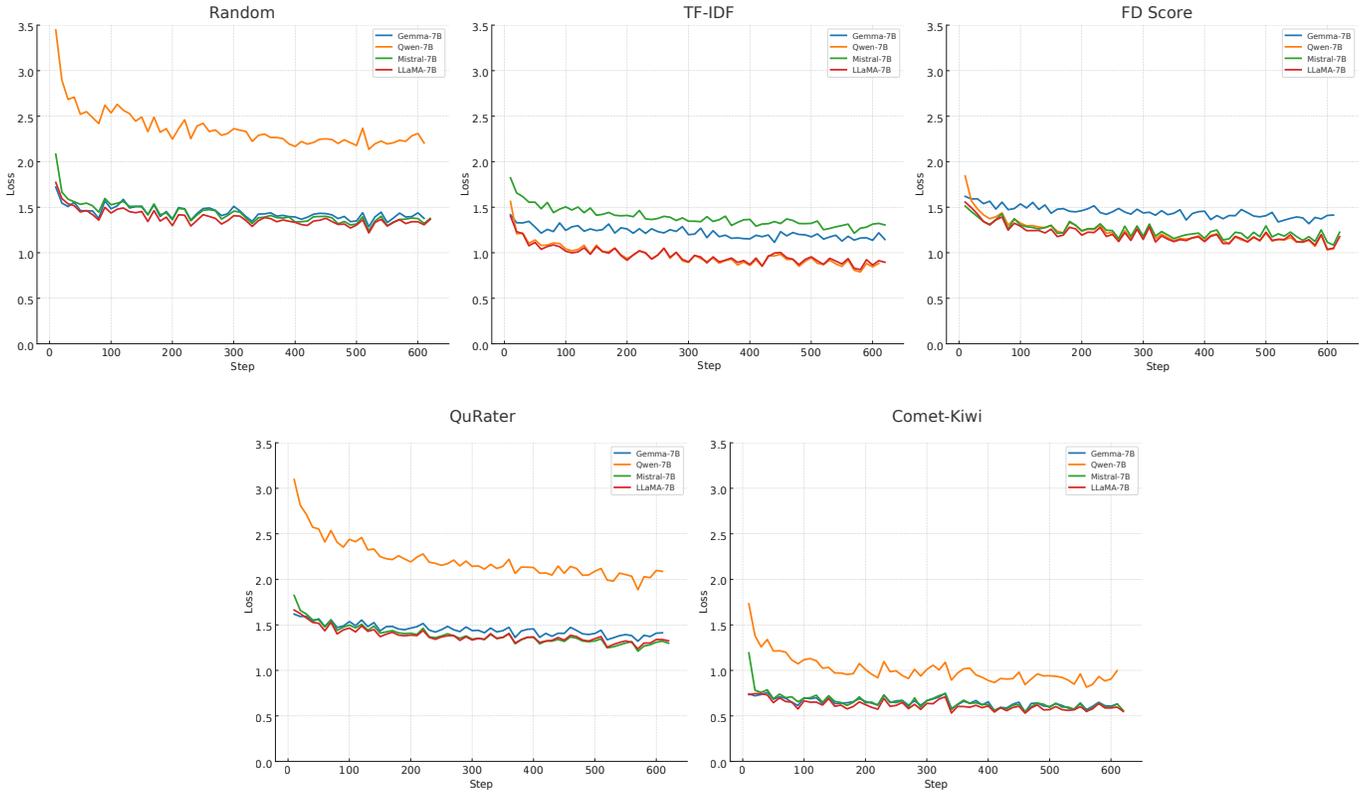}
    \caption{
        Training loss curves for all evaluated models (LLaMA-7B, Gemma-7B, Mistral-7B, and Qwen-7B) fine-tuned on JA$\rightarrow$EN translation. 
        Each subplot corresponds to a different data selection strategy (Random, TF--IDF, FD Score, QuRater, and Comet--Kiwi), enabling a direct comparison of model behavior under consistent training conditions.
    }
    \label{fig:training_loss_comparison_all}
\end{figure*}

Figure~\ref{fig:training_loss_comparison_all} compares the training loss curves on the Japanese$\rightarrow$English (JA→EN) translation task. 

Across all architectures, the COMET-Kiwi method consistently yields the lowest training loss and the smoothest convergence. This result highlights the effectiveness of semantic-based data selection, which prioritizes pairs that are contextually aligned with human judgments of translation quality. 

Furthermore, the similarity of convergence patterns across different model architectures indicates that the observed gains are not model-specific but rather a function of the underlying data distribution. 

The smoother loss trajectories under COMET--Kiwi reflect more coherent learning dynamics, likely because the model encounters fewer noisy or inconsistent training pairs. These findings reinforce the hypothesis that \textit{data quality, rather than data quantity}, is the dominant factor driving fine-tuning performance in low-resource settings. 

In other words, selecting fewer but semantically consistent samples can yield more stable training and better generalization than using larger, lexically diverse but noisier datasets.

\section{Conclusion}

This study investigated how different data selection strategies affect the fine-tuning quality of large language models for Japanese–English translation. 
Through extensive experiments across multiple 7B-scale architectures, we demonstrated that \textbf{semantic-based selectors}, particularly \textbf{COMET–Kiwi}, consistently outperform lexical or geometry-based heuristics such as TF–IDF and FD–Score. 

Our analyses further revealed that semantic selectors not only improve mean \comet{} and \bleu{} scores but also reduce score variance, suggesting greater stability and representational coherence in the selected subsets. 
Qualitative inspection confirmed that COMET–Kiwi–filtered samples produce idiomatic, contextually faithful, and stylistically consistent outputs—characteristics that lexical heuristics often fail to capture.  

These findings underscore a central insight: \emph{data quality, not data quantity, is the key driver of effective LLM fine-tuning under limited data sizes}. 
By selecting fewer but semantically richer examples, models learn more robust cross-lingual mappings and converge more smoothly during training.  

Future work will explore hybrid selection frameworks or predictive utility models. 
We also plan to extend this approach to other language pairs and downstream tasks to further validate the scalability of semantic-aware filtering in low-resource adaptation scenarios. 
Ultimately, this research highlights the importance of integrating learned quality estimation into modern LLM pipelines for more efficient, human-aligned machine translation.

\section{Limitations}

Our study is limited to Japanese$\leftrightarrow$English translation in a deliberately low-resource setting. 
This choice reflects our primary goal of understanding how data selection behaves when fine-tuning budgets are tight, rather than benchmarking absolute state-of-the-art performance. 
As a result, it remains to be seen whether the relative advantages of semantic selection methods extend to other language pairs, domains, or data scales.

In addition, we use a single fine-tuning configuration and evaluate models only with automatic metrics (\bleu{} and \comet{}). 
Varying optimization hyperparameters (e.g., learning rate, batch size, number of updates), fine-tuning set sizes, or mixing multiple domains could further reveal how robust each selection method is under different training dynamics. 
Likewise, complementing automatic scores with targeted human evaluation would provide a more complete picture of how semantic selection impacts perceived translation quality.

\end{document}